\documentclass{ifacconf}

\usepackage{graphicx}
\usepackage{natbib}
\usepackage{multirow}
\usepackage{amsmath}
\begin{document}
\begin{frontmatter}

\title{Zero-Shot Learning for Obsolescence Risk Forecasting\thanksref{footnoteinfo}} 

\thanks[footnoteinfo]{This work was supported by SNCF RESEAU.}

\author[supmeca,sncf]{Elie Saad} 
\author[supmeca]{Aya Mrabah} 
\author[supmeca]{Mariem Besbes}
\author[supmeca]{Marc Zolghadri}
\author[sncf]{Victor Czmil}
\author[laas]{Claude Baron}
\author[sncf]{Vincent Bourgeois}

\address[supmeca]{ISAE-Supméca, 3 Rue Fernand Hainaut, 93400 Saint-Ouen-sur-Seine, France (e-mail: \{elie.saad,aya.mrabah,mariem.besbes,marc.zolghadri\}@isae-supmeca.fr).}
\address[sncf]{SNCF - Campus Fruitier (Eurostade), 6 Av. François Mitterrand, 93210 Saint-Denis, France (e-mail: \{elie.saad,victor.czmil,vincent.bourgeois1\}@reseau.sncf.fr)}
\address[laas]{LAAS-CNRS, 7 Av du Colonel Roche, 31400, Toulouse, France (e-mail: claude.baron@insa-toulouse.fr)}

\begin{abstract}
Component obsolescence poses significant challenges in industries reliant on electronic components, causing increased costs and disruptions in the security and availability of systems. Accurate obsolescence risk prediction is essential but hindered by a lack of reliable data. This paper proposes a novel approach to forecasting obsolescence risk using zero-shot learning (ZSL) with large language models (LLMs) to address data limitations by leveraging domain-specific knowledge from tabular datasets. Applied to two real-world datasets, the method demonstrates effective risk prediction. A comparative evaluation of four LLMs underscores the importance of selecting the right model for specific forecasting tasks.
\end{abstract}


\end{frontmatter}
%===============================================================================

\section{Introduction}

Obsolescence is a significant challenge for industries relying on electronic components and complex systems. It occurs when products become outdated or unavailable due to technological advancements, market changes, or new regulations \citep{IEC62402_2019}, leading to increased maintenance costs, disruptions in the security and availability of systems \citep{zolghadri2023micro}, and operational inefficiencies \citep{Mellal_2020}.

Proactive obsolescence management is a strategic approach to address component obsolescence in long-lifecycle systems, particularly in complex product systems \citep{meng2014strategic}. It involves monitoring and prioritizing the entire product structure to provide near- and mid-term solutions, e.g., through obsolescence risk forecasting. Forecasting obsolescence risk has become crucial due to rapid technological advancements, especially for long-life systems \citep{jennings2016forecasting}. Traditional methods often rely on subjective inputs and laborious market analyses, leading to inconsistent predictions. To overcome these challenges, machine learning-based approaches have been developed for more accurate and efficient obsolescence forecasting \citep{jennings2016forecasting}.

There is a critical need for improved forecasting techniques and proactive management to address the challenges of component obsolescence \citep{rust2022literature}. A significant issue in obsolescence forecasting, particularly for electronic components, is the lack of sufficient and reliable data as companies may not consistently document obsolescence issues \citep{Gupta_2021}. This includes problems such as incomplete datasets and difficulties in accessing relevant information \citep{zolghadri2023micro}. Additionally, the limitations of available data further constrain the effectiveness of forecasting efforts \citep{moon2022forecasting}.

Zero-shot learning (ZSL) offers a promising solution to the challenge of limited labeled data, which refers to datasets where each instance is annotated with specific information or categories required for training machine learning models, in various domains. It is a machine learning approach that can classify data into categories not seen during training, reducing the need for available data \citep{riordan2024apply}. ZSL is a specialized form of transfer learning \citep{lazaros2024comprehensive}, i.e., using knowledge gained from one task to improve learning in a related, new task. Tabular datasets, which are rich in domain-specific information, pose challenges for straightforward transfer learning \citep{wang2024data}, requiring customized approaches and the incorporation of domain knowledge for effective predictions. To solve this, recent researches have explored ZSL for tabular data classification using large language models (LLMs), where the approaches leverage prior knowledge encoded in LLMs \citep{mandal2024initial,sui2024table}. Despite these advancements, research suggests varying results on different tasks \citep{hegselmann2023tabllm}.

Inspired by these developments, this paper investigates how recent LLMs can be utilized to forecast obsolescence risk using tabular data. The effectiveness of these LLMs is assessed through two use-case datasets: a systems dataset and an electronic components dataset. To the best of our knowledge, no prior work has addressed ZSL for obsolescence prediction. The contributions of this work are twofold:
\begin{itemize}
    \item Introducing a novel approach for forecasting the state of parts without requiring prior availability of data.
    \item Conducting a comparative evaluation of recently released LLMs in a zero-shot learning setting.
\end{itemize}

The code and datasets utilized in this study are publicly available on GitHub\footnote{https://github.com/Inars/Zero-Shot-Learning-for-Obsolescence-Risk-Forecasting}, ensuring reproducibility and supporting further advancements in the field.

The rest of the paper is organized as follows: Section~\ref{sec:literature_review} reviews existing obsolescence risk forecasting methods. Section~\ref{sec:methodology} explains the methodology. In Section~\ref{sec:experimentation}, experimental results are presented and model performances are discussed. Finally, Section~\ref{sec:conclusion} concludes the paper and suggests directions for future research.
\section{Literature Review}\label{sec:literature_review}
The field of obsolescence risk forecasting has seen growing emphasis on machine learning techniques, particularly for predicting the likelihood of a part becoming obsolete. Supervised learning is often favored over unsupervised methods, as highlighted by \cite{jennings2016forecasting}, due to its ability to leverage known data states and avoid ambiguous clustering outcomes. Their study, which used a dataset scraped from the GSM Arena website, identified random forest as the most effective algorithm for obsolescence risk forecasting. This conclusion was reinforced by \cite{grichi2017random}, who argued that random forests' resilience against dataset variance and bias makes it ideal for such tasks. Building on this, \cite{grichi2018optimization} and \cite{grichi2018new} integrated meta-heuristic genetic and particle swarm optimization algorithms into the random forest framework, enhancing its predictive performance.

Further support for random forest comes from \cite{trabelsi2021obsolescence}, who evaluated multiple feature selection approaches (filter, wrapper, and embedded) across five machine learning algorithms, including artificial neural networks and support vector machines. Their findings echoed earlier research, affirming that random forests achieved the highest precision in predicting obsolescence risk. Like the previous studies, their analysis also relied on technical data scraped from GSM Arena.

An alternative two-stage forecasting method is proposed by \cite{liu2022obsolescence}. This approach combines the ELECTRE I method for feature selection and an enhanced radial basis function neural network for prediction. Innovations such as improved particle swarm optimization, gradient descent enhancements, and data weighting techniques significantly improved clustering, convergence speed, and prediction accuracy.

While the existing methods discussed in the literature demonstrate significant advancements in obsolescence risk forecasting, they all rely on substantial amounts of available data for training. This dependence on extensive datasets can be a limitation in scenarios where data is incomplete, inaccessible, or unavailable. The approach proposed in this paper addresses this gap by introducing a novel capability to predict the state of parts without requiring prior access to labeled data.
\section{Methodology}\label{sec:methodology}
This section outlines the methodology used to evaluate the capabilities LLMs in forecasting the availability or obsolescence of systems and components. It describes the framework employed in Section~\ref{sec:framework}. The remainder of the section introduces the models used for comparison in Section~\ref{sec:models}, the datasets involved in Section~\ref{sec:datasets}, and the metrics used to assess performance in Section~\ref{sec:metrics}.
\subsection{Framework}\label{sec:framework}
The approach used in this body of work can be summarized as prompting an LLM with various features of a system or an electronic component, and then asking it whether that system or component is available or obsolete. This method is a slightly modified version of the framework introduced by \cite{hegselmann2023tabllm}. The approach is outlined as follows:

\begin{enumerate}
    \item \textbf{Serialization:} The tabular data is serialized into a natural language string representation. Specifically, the "Text Template" serialization method is employed, where each feature is represented textually in the format: ``The \emph{column name} is \emph{value}.''
    \item \textbf{Prompting the LLM:} The serialized data is then used to prompt the LLM, accompanied by a brief description of the classification task. The prompt instructs the LLM to analyze the provided data and predict class labels, i.e., available or obsolete, for new data points. The prompt follows this structure: ``Diode/Phone features: \emph{serialization}. Question: Is this diode/phone available? Yes or no? Answer:'', where ``diode'' is used for the Arrow dataset and ``phone'' for the GSM Arena dataset.
    \item \textbf{Evaluation:} The performance of the LLM-based classifier is evaluated on the full Arrow and GSM Arena datasets, without any fine-tuning.
\end{enumerate}

\subsection{Models}\label{sec:models}
To identify the best ZSL model for forecasting the state of systems and components, four recent notable models were selected for evaluation: Task Zero (T$0$) by Hugging Face, Llama $3$ by Meta, Gemma $2$ by Google, and Phi $3$ by Microsoft.

The first model, T$0$, was introduced by  \cite{sanh2022multitask}. The authors examine the potential of multitask learning in enabling LLMs to generalize across a wide range of tasks without requiring task-specific training, meaning the model does not need to be fine-tuned or trained on data tailored to a specific problem or application. This approach leverages a pretrained autoencoder model that is fine-tuned on a diverse set of datasets, which are reformulated into natural language prompts. The prompts provide task-specific instructions in a human-readable format, enabling the model to interpret and respond to various tasks, even those it has not encountered before. The architecture of the model consists of an encoder-decoder framework, where the encoder processes the input prompts, and the decoder generates the corresponding output, facilitating the ability of the model to adapt to new tasks. By fine-tuning on this multitask mixture, the model achieves strong zero-shot performance, often surpassing much larger models on standard benchmarks. This capability is particularly relevant for predicting the state of parts, as the model can use generalizable knowledge from diverse tasks to understand and forecast obsolescence risk without requiring extensive labeled datasets specific to this domain.

The second model, Llama $3$, introduced by \cite{dubey2024llama}, represents a new set of LLMs designed to handle tasks in multiple languages, coding, reasoning, and tool usage. The models are built on a transformer architecture, a neural network design that uses self-attention mechanisms to process input sequences efficiently by assigning varying levels of importance to different parts of the data. This architecture enables the model to understand relationships within sequences, regardless of their position. The largest version of Llama $3$ contains $3.21$ billion parameters and supports a context window of up to $128$K tokens, where tokens represent the smallest units of text---such as words, subwords, or characters---processed by the model. These tokens are fundamental for breaking down language into manageable units for computation, allowing the model to generate coherent and contextually appropriate responses. The Llama $3$ models are trained in two stages: first, through pre-training on a diverse and clean corpus of text to understand language structure and gain knowledge, and second, via post-training to fine-tune the models for instruction-following behavior. This instruction-following capability is particularly beneficial for predicting the state of a part, as it allows the model to interpret prompts related to obsolescence risk and leverage its extensive pre-trained knowledge to generate accurate predictions, even in the absence of domain-specific training data.

The third model used is the Gemma $2$ model introduced by \cite{team2024gemma}. The Gemma $2$ model family introduces a series of lightweight, open language models designed to balance high performance with practical deployment constraints. These models, ranging from $2$ billion to $27$ billion parameters, leverage key innovations such as knowledge distillation, where smaller models are trained to replicate the performance of larger models, resulting in enhanced efficiency and effectiveness. This technique enables Gemma $2$ models to achieve competitive performance relative to models $2$ to $3$ times their size. Additionally, the models incorporate architectural improvements, including grouped-query attention and local-global attention strategies, which optimize the processing of information. These features make the Gemma $2$ models particularly well-suited for predicting the state of a part, as their efficiency and advanced attention mechanisms allow for the rapid analysis of tabular data and the identification of key patterns indicative of obsolescence risk.

Finally, the Phi $3$ model, detailed by \cite{abdin2024phi}, introduces phi-$3$-mini, a highly efficient $3.8$ billion parameter language model that outperforms much larger models like Mixtral $8$x$7$B \citep{jiang2024mixtral} and GPT-$3.5$, despite its smaller size. The key novelty lies not in increasing the model size, but in enhancing the training dataset, which combines heavily filtered web data with synthetic data generated by language models. The architecture of the model is based on a transformer decoder with $3072$ hidden dimensions, $32$ attention heads, and $32$ layers. It supports a context length of $4$K tokens and utilizes a tokenizer with a vocabulary size of $320,641$. This design prioritizes computational efficiency and robust generalization, making Phi $3$ particularly well-suited for predicting the state of a part. Its reliance on high-quality data allows it to identify subtle patterns in obsolescence risk, even when operating with constrained computational resources, providing an effective and scalable solution for real-world forecasting challenges.
\subsection{Datasets}\label{sec:datasets}
To evaluate the proposed methodology, two distinct case studies were analyzed. The first involves a high-level study on system obsolescence using the GSM Arena dataset, which focuses on smartphone obsolescence as highlighted by \cite{jennings2016forecasting}. The second case study is a component-level obsolescence using the Arrow dataset, which provides data on the Zener diode, an electronic semiconductor component. Information on the Zener diode was sourced from the Arrow \cite{arrow2024} website, a company that distributed electronic components.

\begin{table}[hb]
\begin{center}
\caption{Use case dataset specifications}\label{tb:datasets}
\begin{tabular}{|c|c|c|c|c|}
\hline
\textbf{Dataset} & $\boldsymbol{N}$ & $\boldsymbol{N_0}$ & $\boldsymbol{N_1}$ & \textbf{\textbackslash} \\ \hline
Arrow & $11080$ & $7580$ & $3500$ & $68.41$ \\ \hline
GSM Arena & $8628$ & $4773$ & $3855$ & $55.32$ \\ \hline
\end{tabular}
\end{center}
\end{table}

The various specifications of the two datasets are represented in Table \ref{tb:datasets}. The column $N$ represents the total number of rows in the datasets. The columns $N_0$ and $N_1$ represent the total number of obsolete and available instances respectively. The backslash (\textbackslash) column represents the percentage proportion of obsolete instances within the datasets. 

The Arrow dataset consists of $11,080$ instances, with $7,580$ instances labeled as obsolete and $3,500$ as available, resulting in a proportion of $68.41\%$ obsolete instances. On the other hand, the GSM Arena dataset contains $8,628$ instances, with$ 4,773$ marked as obsolete and $3,855$ as available, giving a proportion of $55.32\%$ for obsolete instances. Both datasets show an imbalance between the number of obsolete and available instances. In particular, the Arrow dataset is highly imbalanced, with a larger proportion of obsolete instances (68.41\%), while the GSM Arena dataset, although somewhat less imbalanced, still leans toward a higher proportion of obsolete instances. However, since this study employs a zero-shot learning setting, where the model makes predictions without being trained on labeled instances from these datasets, the high proportion of obsolete cases does not influence the results of the model. This ensures that the evaluation is unaffected by the data imbalance during the inference stage.
\subsection{Metrics}\label{sec:metrics}
To examine the performance of the four LLM models, the following five evaluation metrics are employed \citep{obi2023comparative}:

\begin{itemize}
    \item \textbf{Accuracy:} Accuracy measures the proportion of correctly classified instances (true positives and true negatives) out of the total instances:
    \begin{equation}
        \text{Accuracy}=\frac{\text{TP}+\text{TN}}{\text{TP}+\text{TN}+\text{FP}+\text{FN}},
    \end{equation}
    where TP (True Positive) is the correctly predicted positive instances. TN (True Negative) is the correctly predicted negative instances. FP (False Positive) is the negative instances incorrectly classified as positive. And FN (False Negative) is the positive instances incorrectly classified as negative. Accuracy ranges from $0$ to $1$, with $1$ indicating perfect classification.
    \item \textbf{Precision:} Precision is applied when minimizing false positives (incorrectly predicting a negative instance as positive) is crucial. It ensures that predicted positive instances are highly likely to be truly positive. Precision quantifies the proportion of correctly predicted positive instances out of all instances predicted as positive:
    \begin{equation}
        \text{Precision}=\frac{\text{TP}}{\text{TP}+\text{FP}}.
    \end{equation}
    Precision ranges from $0$ to $1$, with higher values indicating fewer false positives. The ideal value for precision is $1$, meaning that all predicted positive instances are correctly identified as obsolete components, with no false positives.
    \item \textbf{Recall:} Recall, also known as Sensitivity or True Positive Rate (TPR), is used when minimizing false negatives (failing to identify positive instances) is essential. This metric focuses on capturing all relevant positive instances, even if it leads to some false positives. Recall measures the proportion of correctly predicted positive instances out of all actual positive instances:
    \begin{equation}
        \text{Recall}=\frac{\text{TP}}{\text{TP}+\text{FN}}.
    \end{equation}
    Recall ranges from $0$ to $1$, with higher values reflecting fewer false negatives. The ideal value for recall is $1$, which would indicate that all obsolete components are correctly identified, with no false negatives.
    \item \textbf{F$1$ Score:} The F$1$ score is used when a balance between precision and recall is required. It combines both metrics into a single value, representing a trade-off between the ability to correctly identify positives and the ability to detect all positives. The F$1$ score is the harmonic mean of precision and recall:
    \begin{equation}
        \text{F1}=2\cdot\frac{\text{Precision}\cdot\text{Recall}}{\text{Precision}+\text{Recall}}
    \end{equation}
    The F$1$ score ranges from $0$ to $1$, with $1$ indicating perfect precision and recall. The ideal value for the F$1$ score is $1$, indicating that the model has both high precision and recall, with few false positives and false negatives.
    \item \textbf{Area Under the Curve:} The Area Under the Curve (AUC) metric evaluates the capability of a classifier to distinguish between classes, based on the Receiver Operating Characteristic (ROC) curve, which plots the TPR against the False Positive Rate (FPR) at various threshold settings:
    \begin{equation}
        \text{FPR}=\frac{\text{FP}}{\text{FP}+\text{TN}}.
    \end{equation}
    The AUC represents the area under the ROC curve:
    \begin{equation}
        \text{AUC}=\int_0^1\text{TPR}(\text{FPR})d\text{FPR}
    \end{equation}
    AUC ranges from $0.5$ (random guessing) to $1$ (perfect discrimination). Values above $0.9$ indicate strong classification performance, while values between $0.7$ and $0.9$ suggest moderate performance. The ideal value for AUC is $1$, which represents perfect classification ability with no false positives or false negatives at any threshold.
\end{itemize}

The best model for each dataset is determined through a simple voting technique, where the model with the highest performance across multiple metrics is selected. For the GSM Arena dataset, the model with the most favorable overall performance is chosen, while for the Arrow dataset, a separate model is identified as the best based on similar criteria. This approach ensures the selection of the most suitable model for each dataset independently.
\section{Experimentation}\label{sec:experimentation}
Recent studies highlight that running LLMs demands significant computational resources, including substantial GPU memory and tensor offloading capabilities \citep{isaev2023scaling, borzunov2024distributed}. A tensor is a multi-dimensional array (vector) used to represent data in deep learning models. It generalizes matrices ($2$ dimensional arrays) to higher dimensions. Despite the computational demands of large models, sub-$10$-billion-parameter LLMs present a practical alternative, effectively balancing computational efficiency and accuracy, making them suitable for smaller devices \citep{li2024large}. To address accessibility constraints, this study utilizes small, instruction-tuned variants of the selected models (models fine-tuned for instruction-following behavior), where available. Specifically, the chosen models include T$0$\_$3$B ($2.85$ billion parameters), Llama-$3.2$-$3$B-Instruct ($3.21$ billion parameters), Gemma-$2$-$2$B-It ($2.61$ billion parameters), and Phi-$3.5$-Mini-Instruct ($3.82$ billion parameters).

During the experiments, the models T$0$, Llama $3.2$, and Gemma $2$ successfully adhered to the provided instructions and generated predictions without issues. In contrast, Phi $3.5$ consistently failed to follow the instructions in the prompts, even after modifying the prompts and experimenting with various instructional structures. Consequently, using the prompt structure proposed in Section~\ref{sec:methodology}, Phi $3.5$ was unable to provide predictions for all prompts in the Arrow dataset, missing $276$ predictions out of $11,080$, and failed entirely for the GSM Arena dataset, making no predictions ($0$ out of $8,628$). The primary reason for these failures appears to be the inability of the model to recognize sufficient information in the prompt to make a confident prediction.

\begin{figure}
\begin{center}
\includegraphics[width=8.4cm]{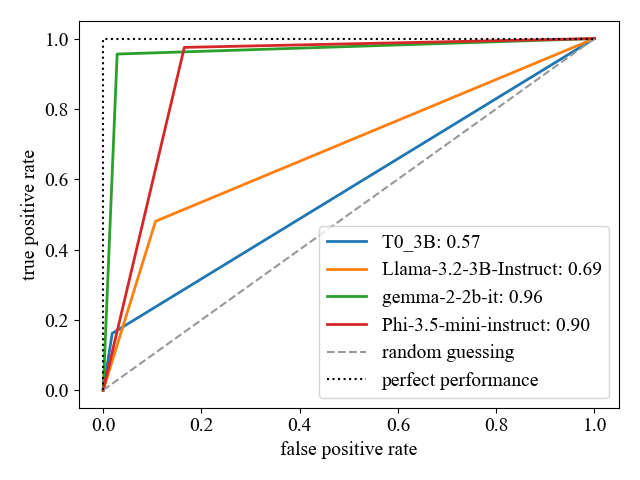}    % The printed column width is 8.4 cm.
\caption{Receiver Operator Characteristic for the Arrow dataset} 
\label{fig:arrow_roc}
\end{center}
\end{figure}

\begin{figure}
\begin{center}
\includegraphics[width=8.4cm]{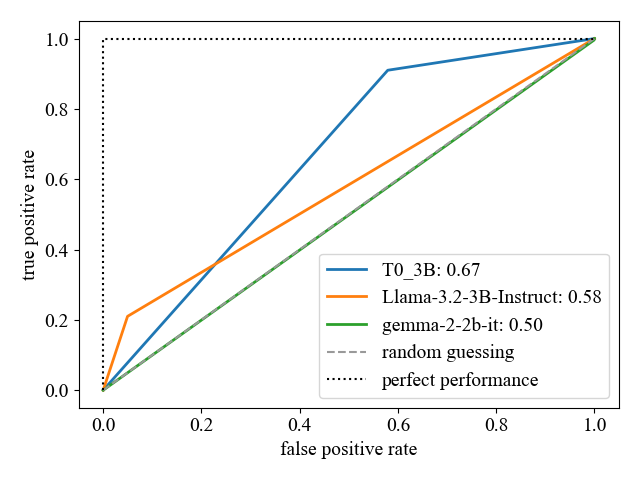}    % The printed column width is 8.4 cm.
\caption{Receiver Operator Characteristic for the GSM Arena dataset} 
\label{fig:phone_roc}
\end{center}
\end{figure}

The experimental results in Table~\ref{tb:results} showcase the performance of the four chosen LLMs applied to the two use-case datasets. Figures \ref{fig:arrow_roc} and \ref{fig:phone_roc} show the ROC curves of the LLMs for the Arrow and GSM Arena datasets, respectively. In this study, TPs refer to instances where the model correctly predicts that a part is available, while TNs refer to instances where the model correctly predicts that a component is obsolete.

\begin{table}[hb]
\begin{center}
\caption{Experimental results}\label{tb:results}
\begin{tabular}{|c|c|c|c|}
\hline
\textbf{Model} & \textbf{Metric} & \textbf{Arrow} & \textbf{GSM Arena} \\ \hline
\multirow{3}{*}{T$0$} & Accuracy & $72.26$ & $\boldsymbol{69.14}$ \\ \cline{2-4}
                       & Precision & $0.803$ & $0.66$ \\ \cline{2-4}
                       & Recall & $0.161$ & $0.91$ \\ \cline{2-4}
                       & F$1$ & $0.269$ & $\boldsymbol{0.765}$ \\ \cline{2-4}
                       & AUC & $0.572$ & $\boldsymbol{0.665}$ \\ \hline
\multirow{3}{*}{Llama 3.2} & Accuracy & $76.3$ & $54.08$ \\ \cline{2-4}
                       & Precision  & $0.676$ & $\boldsymbol{0.84}$ \\ \cline{2-4}
                       & Recall & $0.48$ & $0.21$ \\ \cline{2-4}
                       & F$1$ & $0.561$ & $0.336$ \\ \cline{2-4}
                       & AUC & $0.687$ & $0.58$ \\ \hline
\multirow{3}{*}{Gemma 2} & Accuracy & $\boldsymbol{96.67}$ & $55.12$ \\ \cline{2-4}
                       & Precision  & $\boldsymbol{0.94}$ & $0.552$ \\ \cline{2-4}
                       & Recall & $\boldsymbol{0.956}$ & $\boldsymbol{0.996}$ \\ \cline{2-4}
                       & F$1$ & $\boldsymbol{0.948}$ & $0.711$ \\ \cline{2-4}
                       & AUC & $\boldsymbol{0.964}$ & $0.498$ \\ \hline
\multirow{3}{*}{Phi 3.5} & Accuracy & $72.26$ & --- \\ \cline{2-4}
                       & Precision & $0.803$ & --- \\ \cline{2-4}
                       & Recall & $0.161$ & --- \\ \cline{2-4}
                       & F$1$ & $0.269$ & --- \\ \cline{2-4}
                       & AUC & $0.572$ & --- \\ \hline
\end{tabular}
\end{center}
\end{table}

For the Arrow dataset, Gemma $2$ outperforms all other models across several metrics, achieving an accuracy of $96.67\%$, a precision of $0.94$, and a recall of $0.956$. It also leads with a high AUC of $0.964$, indicating its strong classification capabilities. The T$0$ model, while showing decent performance with an accuracy of $72.26\%$, falls short in recall ($0.161$), suggesting a higher rate of false negatives in its predictions. The Llama $3.2$ model offers better precision ($0.676$) but has a lower recall ($0.48$), indicating a tendency to miss many positive instances. The Phi $3.5$ model, with the same performance as T$0$ in the Arrow dataset, shows identical precision and recall but a much lower AUC ($0.572$), pointing to its limited ability to distinguish between classes.

For the GSM Arena dataset, T$0$ leads in accuracy ($69.14\%$) and the F$1$ score ($0.765$), suggesting it strikes a good balance between precision and recall. However, Llama $3.2$ excels in precision ($0.84$) but suffers from lower recall ($0.21$), which implies that it correctly identifies many positives but also misses a significant portion of them. Gemma $2$, while performing well in recall ($0.996$), does not achieve competitive accuracy or AUC values, reflecting a possible trade-off between sensitivity and overall classification ability.

Using the voting technique outlined in Section~\ref{sec:methodology}, the optimal model for each dataset, and consequently each task, is identified based on performance across multiple metrics. For the Arrow dataset, Gemma $2$ demonstrates superior performance, making it the most suitable choice for forecasting the state of electronic components. Conversely, for the GSM Arena dataset, the T$0$ model emerges as the best-performing model, making it the preferred option for system state forecasting tasks, although its precision is not as significant.
\section{Conclusion}\label{sec:conclusion}
This paper has presented a novel approach for forecasting component obsolescence risk using ZSL with LLMs. By leveraging domain-specific knowledge from tabular datasets, this paper have demonstrated that LLMs can effectively address the challenges of limited labeled data in obsolescence forecasting. Through the application of this method to two real-world datasets, one involving systems and the other focusing on electronic components, the results highlight the potential of ZSL to enhance obsolescence risk prediction accuracy without the need for extensive data collection or prior training on labeled instances.

The comparative evaluation of four recent LLMs in the context of ZSL demonstrated varying performance levels, highlighting the importance of selecting the appropriate model for specific obsolescence forecasting tasks. Notably, the Gemma $2$ model by Google outperformed its counterparts on the Arrow dataset, while the T$0$ model by Hugging Face excelled on the GSM Arena dataset.

Future research in this area should investigate the potential of fine-tuning these models for the specific tasks of systems and electronic component obsolescence prediction within a few-shot learning framework. Additionally, exploring hybrid approaches that integrate LLMs with traditional obsolescence prediction methods could further enhance the predictive accuracy and robustness of obsolescence risk forecasting.
\bibliography{ifacconf}
\end{document}